\documentclass{article}

\usepackage{graphicx} 
\usepackage{wrapfig}
\usepackage{placeins}
\usepackage[symbol]{footmisc}
\usepackage{float} 
\usepackage[colorlinks=true, linkcolor=blue, urlcolor=blue, citecolor=black ]{hyperref}

\usepackage[final]{neurips_2025}

\usepackage[utf8]{inputenc} 
\usepackage[T1]{fontenc}    
\usepackage{amssymb} 
\usepackage{url}            
\usepackage{booktabs}       
\usepackage{amsfonts}       
\usepackage{nicefrac}       
\usepackage{microtype}      
\usepackage{xcolor}         
\usepackage{amsmath}
\usepackage{fontawesome5} 
\newcommand{\emailsymbol}{\faEnvelope} 
\usepackage{marvosym}

\title{Don’t Think of the White Bear: Ironic Negation in Transformer Models Under Cognitive Load}

\author{
Logan Mann $^{\text{\tiny{\emailsymbol}}1*}$ 
  \And
  Nayan Saxena$^{2*}$
  \AND
   Sarah Tandon$^{3*}$ 
  \And
  Chenhao Sun$^{4*}$ 
  \And
  Savar Toteja$^{5*}$ 
  \And
  Kevin Zhu$^{6}$
}

\begin{document}
\maketitle
\begin{abstract}
Negation instructions such as 'do not mention $X$' can paradoxically increase the accessibility of $X$ in human thought, a phenomenon known as ironic rebound. Large language models (LLMs) face the same challenge: suppressing a concept requires internally activating it, which may prime rebound instead of avoidance. We investigated this tension with two experiments. \textbf{(1) Load \& content}: after a negation instruction, we vary distractor text (semantic, syntactic, repetition) and measure rebound strength. \textbf{(2) Polarity separation}: We test whether models distinguish neutral from negative framings of the same concept and whether this separation predicts rebound persistence. Results show that rebound consistently arises immediately after negation and intensifies with longer or semantic distractors, while repetition supports suppression. Stronger polarity separation correlates with more persistent rebound. Together, these findings, complemented by a circuit tracing analysis that identifies sparse middle-layer attention heads amplifying forbidden tokens while early layers suppress, link cognitive predictions of ironic rebound with mechanistic insights into long-context interference. To support future work, we release ReboundBench, a dataset of $5,000$ systematically varied negation prompts designed to probe rebound in LLMs.
\end{abstract}

\section{Introduction}
\footnotetext{$^{*}$Equal contribution.  
$^{1}$University of California, Santa Barbara , $^{2}$Project Lead,  $^{3}$Duke University $^{4}$University of Toronto,  $^{5}$University of Maryland, College Park, $^{6}$Algoverse AI Research 
$\text{\tiny{\emailsymbol}} : $ \texttt{loganmann@ucsb.edu}}

Human attempts to suppress thoughts can paradoxically increase their accessibility, a phenomenon known as \textbf{ironic rebound} \citep{wegner1993ironic,wang2020ironic,hwang2024not}. Classic demonstrations (e.g., "do not think of a white bear") show negation may highlight the forbidden concept \citep{wegner1987paradoxical}. Cognitive accounts attribute this effect to semantic priming and psychological reactance: monitoring processes bring the target concept to mind to check for compliance, and restrictions on thought can provoke a counteractive focus \citep{skenderi2017semantic,miron2006reactance,steindl2015understanding,rosenberg201850}. These findings suggest a general principle: suppression under load can backfire \citep{wegner1993ironic}.

Large language models (LLMs) face a similar interpretive challenge. To follow an instruction like “do not mention $X$,” they activate an internal representation of $X$. Intuitively, this activation could produce rebound-like behavior, making the forbidden concept more likely rather than less \citep{ali2024mitigating}. Prior studies of negation in LLMs support this intuition, showing that models occasionally fail to respect negated instructions \citep{vrabcova2025negation,varshney2024investigating}. Such failures are not only theoretically interesting but also practically significant for safety and moderation, where filtering rules rely on negation.

At the same time, mechanistic interpretability work points to architectural forces that may resist rebound. Certain attention heads (e.g., L10H7 in GPT-2 Small) implement a copy suppression mechanism that reduces the probability of tokens that are deemed forbidden (e.g., due to negation instructions)\citep{mcdougall2023copysuppressioncomprehensivelyunderstanding}. Prior work further shows that LLMs often struggle to retrieve information placed in the middle of long contexts - a phenomenon known as lost-in-the-middle \citep{liu2023lost}. These counterforces suggest that, unlike humans, LLMs may display systematic suppression effects \citep{mcdougall2024copy}. This tension raises the question: under what conditions do models show rebound, suppression, or both?

Existing evaluations leave this question unresolved. Prior work often blurs lexical repetition avoidance with genuine semantic negation, and rarely manipulates distractor text beyond trivial additions of text \citep{liu2023lost, ai6010012}. We lack clarity on how intervening load affects negation in LLMs. This gap is consequential: if suppression is fragile to paraphrase or distraction, moderation systems based on simple negation heuristics may be unreliable.

In this paper, we make two main contributions. First, we introduce ReboundBench, a dataset of 5,000 negation prompts with systematically varied distractor types (semantic, syntactic, repetition) and lengths, enabling controlled study of ironic rebound dynamics in LLMs. Second, through two experiments (i) varying distractor load to test when suppression weakens and rebound emerges, and (ii) measuring polarity separation across framings. We show how rebound strength and persistence vary across model families and scales and verify this through circuit tracing analysis. We evaluate nine open-source models ranging from GPT-2 Small to GPT-OSS-20B \citep{Radford2019LanguageMA, biderman2023pythiasuiteanalyzinglarge, workshop2022bloom, zhang2022opt, grattafiori2024llama, yang2025qwen3, black2022gptneox20bopensourceautoregressivelanguage, agarwal2025gpt, liquidai_lfm2_350m}.

\section{Methods}
\subsection{Preliminaries}
\paragraph{Log-probability }
Each model’s native tokenizer is used; when a candidate target concept $X$ tokenizes into multiple subwords, the item is either replaced with a single-token synonym or excluded, depending on availability. We evaluate,
$
    \log p_\theta(X \mid c) \;=\; \log \prod_{t=1}^T p_\theta(x_t \mid c, x_{<t}),
$
where $X=(x_1,\dots,x_T)$ is the tokenized form of the target concept. Since our dataset restricts $X$ to single tokens, this simplifies to a single conditional log-probability. Outliers and skipped items (multi-token $X$ without synonym replacement) are excluded consistently across conditions.

\paragraph{Load}
We implemented load ($\ell$) as the amount of distracting texts that the model must process. To account for different context windows we converted the raw lengths of the distractors to within-model percentiles; $\ell = \text{rank\_pct}(\text{distractor\_length})_{\text{within model}} \in [0,1]$ where $\ell = 0$ represents minimal text between instruction and measurement, and $\ell = 1$ represents maximum text. To properly compare the effects in log-probability described below, we establish a \textbf{high-load baseline}. By calculating the mean log-probability when the cognitive load is the highest (top $5\%$ of $\ell$), denoted by $\mu_{\text{base}} = \mathbb{E}[\log_2 p(\text{forbidden}) \mid \ell \geq 0.95]$. At these points, the negation instruction is maximally attenuated, revealing the model's natural probability for the forbidden token.

\paragraph{Surprisal}
To quantify the ironic rebound effect, we measure the change in the surprisal of the forbidden token relative to a high-load baseline. Surprisal, $s$, measures unexpectedness in bits: $s(x|c) = -\log_2 p_\theta(x|c)$. We define the surprisal difference, $\Delta s(\ell)$, as the change in surprisal at a given load percentile compared to the baseline surprisal under maximum load, $s_{\text{base}} = \mathbb{E}[-\log_2 p(\text{forbidden}) \mid \ell \ge 0.95]$. This difference can be expressed in terms of log-probabilities:
$$ 
\Delta s(\ell) = s_{\text{base}} - s(\text{forbidden} \mid\ell) = \log_2 p(\text{forbidden} \mid \ell) - \mu_{\text{base}} 
$$
Here, $\mu_{\text{base}}$ is the mean base-2 log-probability at maximum load ($\ell \ge 0.95$). Positive values of $\Delta s(\ell)$ (in bits) indicate ironic rebound---the negation instruction has made the forbidden token less surprising (more probable) than it would be at baseline. Negative values indicate suppression.

\paragraph{Suppression score}
We quantify suppression versus rebound with a score that compares the log-probability of the target under the suppression and neutral conditions, such that,
$
S(L) = \mathbb{E}_{(topic,X)} \Big[\log p_\theta\!\left(X \mid c_{\text{neu}}(L)\right) - \log p_\theta\!\left(X \mid c_{\text{sup}}(L)\right)\Big].
$
Here $c_{\text{neu}}(L)$ is the neutral context with load length $L$, and $c_{\text{sup}}(L)$ is the suppression context of the same length. $S(L) < 0$ indicates rebound, where the negation instruction increases the likelihood of $X$.

\paragraph{Polarity discrimination}
To test whether models distinguish between attitudinal framings, we generate two minimal variants of each context: neutral and negative continuations about the same concept $X$ and summarize their separation with the following metric:
\begin{align}
    \Delta = \log p_\theta(r_{\text{neu}}) - \log p_\theta(r_{\text{neg}}).
\end{align}
Larger values of $\Delta$ reflect stronger polarity discrimination, while smaller values indicate that the model assigns similar probabilities regardless of framing.

\paragraph{Statistical model}
At the item level, we fit a mixed-effects regression of surprisal difference in bits:
\begin{align*}
    \Delta(\text{bits}) \sim \ell + \text{type} + \ell:\text{type} + \text{model} + (1 \mid\text{forbidden\_concept}) .
\end{align*}
This specification estimates how rebound strength changes with load, whether collapse dynamics differ by distractor type, and whether effects vary by model family. The random intercept accounts for dependence among items sharing the same forbidden concept. As a robustness check, we also fit cluster-robust OLS regressions. Coefficients of interest are $\beta_{\text{load}}$ (attenuation with load) and $\beta_{\ell \times \text{type}}$ (interaction between load and distractor type).

\paragraph{AUC$_\Delta$}
AUC is the total rebound mass, integrating positive $\Delta$ across the load range
\begin{align*}
    \text{AUC$\Delta$} = \int_{0}^{1} \max(0, \Delta(\ell)) \,d \ell .
\end{align*}

\subsection{Experimental setup}
\paragraph{Models}
We evaluated nine open-source language models spanning different parameter scales and training regimes: GPT-OSS-20B \citep{agarwal2025gpt}, GPT-Neox-20B \citep{black2022gptneox20bopensourceautoregressivelanguage}, Qwen3-14B \citep{yang2025qwen3}, Llama-3-8B-Instruct \citep{grattafiori2024llama}, OPT-2.7B \citep{zhang2022opt}, Bloom-560M \citep{workshop2022bloom},  Pythia-410M \citep{biderman2023pythiasuiteanalyzinglarge}, LFM2-350M \citep{liquidai_lfm2_350m}, and GPT-2 Small \citep{Radford2019LanguageMA}. This selection covers both base and instruction-tuned models, providing a range of capacities for handling negation and semantic distinctions. Models are accessed via the Hugging Face Transformers library with code publicly available at: 
\href{https://github.com/cesium132dot9/Dont-Think-of-the-White-Bear}{\texttt{https://github.com/cesium132dot9/Dont-Think-of-the-White-Bear}}

\paragraph{Dataset}
We introduce \textbf{ReboundBench}, 5{,}000 programmatic, templated prompts, publicly available at  \href{https://huggingface.co/datasets/SavarToteja/dont-think-of-the-white-bear}{Link to Dataset}. Each prompt pairs a topic with a target concept $X$ (topic supplies local discourse context; $X$ is the negated word). The corpus is balanced across conditions and minimizes hand‑selection. All text is lowercased, whitespace‑normalized, and punctuation‑stripped to reduce tokenization artifacts. Target concepts are common nouns and single tokens under the evaluated tokenizers. The same dataset is used for all models and experiments to enable direct comparison. More details about the dataset are in section \ref{sec: dataset details}. 

\paragraph{Evaluation procedure}
Two experiments are implemented via a custom evaluation script.
\begin{itemize}
    \item \textbf{Suppression under load} Each trial begins with an introduction of the topic and a negation instruction of the form ``do not mention $X$.'' An intervening load is then inserted before the evaluation point. Loads vary in both length ($\ell\in\{0,\dots,1024\}$ tokens, sampled in powers of two) and genre: syntactic fillers (grammatical but semantically light continuations), semantic content (coherent text related to the topic but excluding $X$), or repetition (restatements of surrounding content without directly naming $X$). A matched neutral condition omits the negation instruction. We then use the Suppression Score (S(L)) and the Surprisal Difference ($\Delta s(\ell)$) metrics to quantify the results.
    \item \textbf{Polarity discrimination} For each target concept, we constructed two context variants (negative and neutral) and measured the model’s ability to distinguish between them using the Polarity Discrimination ($\Delta$) metric. A high $\Delta$ indicates that the model assigns clearly different probabilities to the three framings, while a low $\Delta$ reflects little separation. As shown in the figures, we plot the distribution of $\Delta$ scores across topic–forbidden pairs for several models to compare their capacity for polarity discrimination.
\end{itemize}

\section{Results}
\begin{figure}[htbp]
    \centering
    \includegraphics[width=1\linewidth]{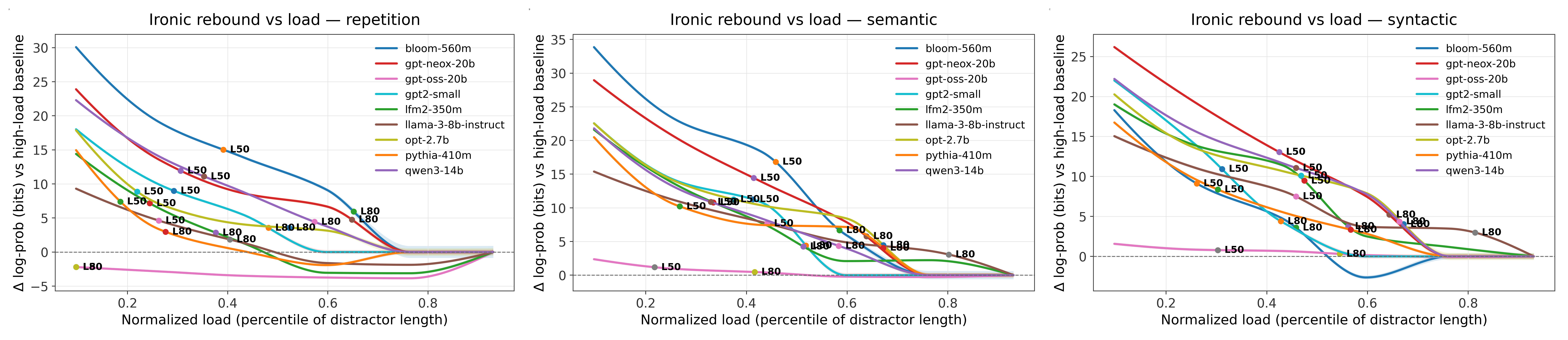}
    \caption{Change in surprisal ($\Delta$ bits) of the forbidden token relative to a high-load baseline as normalized distractor length increases: semantic, syntactic, repetition}
    \label{fig: ironic rebound vs load}
\end{figure}

\paragraph{Suppression \& rebound under load: }
LLMs mimic ironic rebound when told not to produce a concept, as shown in Figure \ref{fig: ironic rebound vs load}. Semantic distractors trigger the strongest rebound, syntactic distractors are weaker, and repetition is the weakest. Smaller models show only brief rebound before suppression, while mid-scale models sustain stronger effects: Bloom remains robust across distractors, Qwen separates semantic from syntactic rebound, and OPT-2.7B combines higher magnitude with persistence. GPT-OSS-20B is an outlier, showing minimal or even negative rebound, breaking the scaling trend. Granular model-level results are provided in the Appendix (Figure \ref{fig:appendix-model-graphs} and Table \ref{tab:rebound_summary}).

\begin{wrapfigure}{R}{0.45\textwidth} 
    \vspace*{-1cm}
    \centering
    \vspace{0.5cm}
    \includegraphics[width=\linewidth]{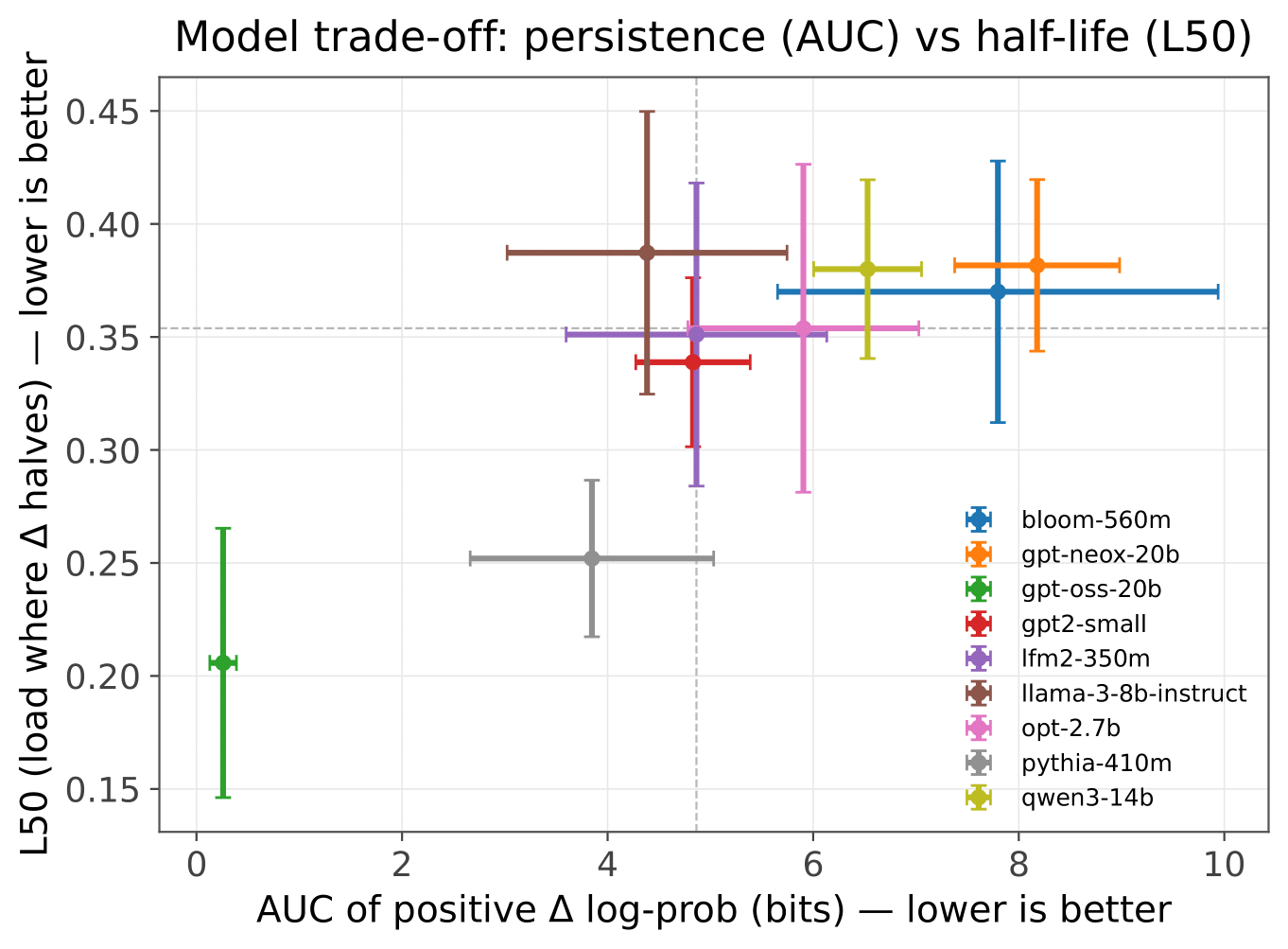}
    \caption{Trade-off between rebound magnitude (AUC) and persistence ($L_{50}$, the load at which rebound halves)}
    \label{fig:auc_trade-off}
\end{wrapfigure}

\paragraph{Polarity discrimination}
Polarity discrimination is not monotonic with scale. Family effects dominate: Llama-3-8B-Instruct (11.3 bits) and Bloom-560M (10.1) show the strongest separation, while larger models such as GPT-NeoX-20B (5.7) and Qwen-14B (6.5) are weaker. GPT-OSS-20B and Pythia-410M are outliers, the weakest by far. Polarity separation also correlates with rebound persistence $L_{50}$ ($r \approx 0.44$), linking sharper attitudinal encoding to longer-lived rebound. This suggests that models with more nuanced semantic representations may struggle more to suppress activated concepts over long contexts, pointing to a potential tension between representational quality and the stability of cognitive control. Figure \ref{fig:auc_trade-off} shows a clear trade-off where models with a larger rebound magnitude have a longer-lived rebound. Higher AUC tends to pair with higher $L_{50}$, so models closer to the lower-left are preferable for negation control.

\FloatBarrier
\paragraph{Internal Mechanisms of Ironic Rebound}
Having established that language models exhibit ironic rebound when told not to mention specific concepts, we next sought to understand what happens inside the models to produce this effect. We focused on attention heads, specialized components within transformers that determine what information to focus on at each processing step. By systematically disabling individual heads and measuring how this changed the model's tendency to produce forbidden words, we could identify which heads were most responsible for ironic rebound.

\begin{figure}[H]
    \centering
    \includegraphics[width=0.9\textwidth]{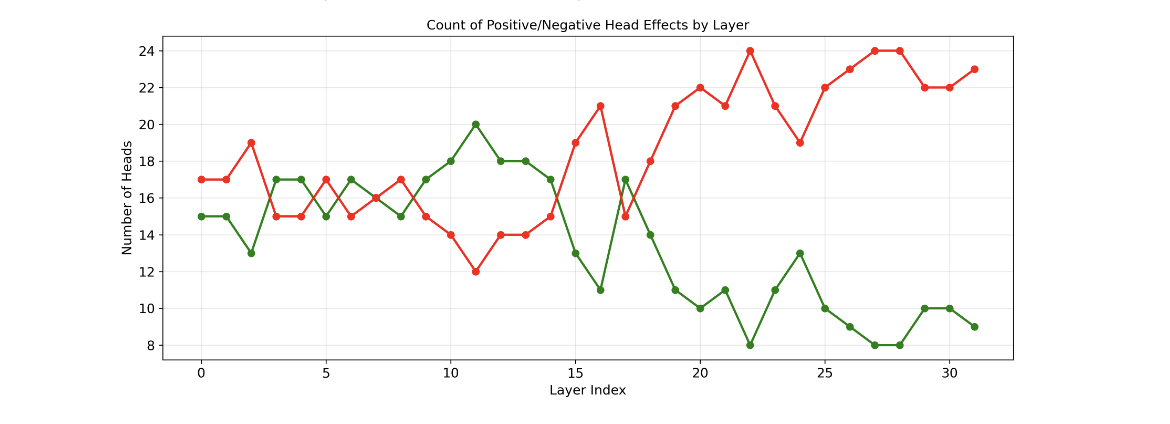}
    \caption{How suppression and amplification effects change across model layers. Early layers suppress, middle layers show mixed effects with emerging rebound, and later layers stabilize.}
    \label{fig:layer_progression}
\end{figure}

\nopagebreak
By examining how suppression and amplification effects change across the model’s depth, we
discovered a clear pattern. Early layers (0–7) tend to suppress forbidden concepts, consistent with
the model initially trying to follow instructions. Middle layers (8–16) show the most variability,
with both strong suppression and rebound effects. Later layers (20+) generally have smaller effects,
suggesting the major decisions about forbidden tokens are made earlier in processing.
Figure \ref{fig:layer_progression} illustrates this progression, showing both the average effect per layer and the count of
suppressive versus amplifying heads. The pattern reveals a kind of internal struggle; the model
initially attempts suppression but then partially reverses course through amplifying heads in middle
layers.
\FloatBarrier 

\begin{figure}[H]
    \centering
    \includegraphics[width=0.5\textwidth]{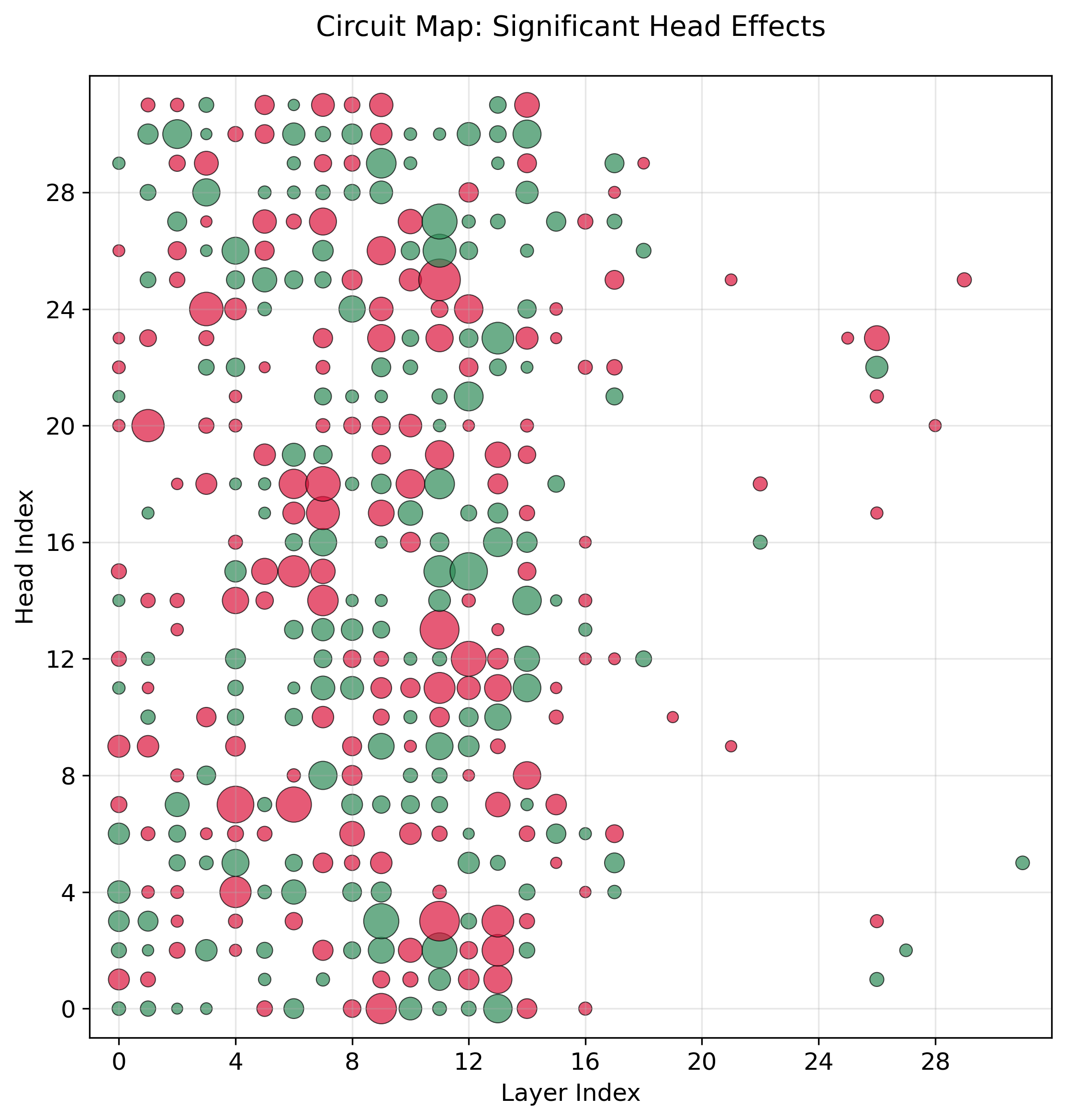}
    \caption{Circuit map of the most influential attention heads. Circle size indicates effect magnitude; color indicates direction (red = amplification, green = suppression). Effects cluster in the middle layers.}
    \vspace{-2pt}
    \label{fig:circuit_map}
\end{figure}

\nopagebreak
Despite involving over 1,000 attention heads, the circuits responsible for ironic rebound are remark-
ably sparse. Typically, just 15–20 heads (roughly 1–2 per layer, concentrated in middle layers)
account for 80\% or more of the total rebound effect. Figure \ref{fig:circuit_map} visualizes these key heads, showing
their positions and effect magnitudes. The strongest suppressors tend to appear around layers 8–13,
while the most potent amplifiers concentrate in layers 12–17.

This localization is practically important; it suggests that ironic rebound could potentially be mitigated
by targeted interventions on specific heads, rather than requiring holistic model changes. Across the full model architecture, 26.7\% of heads contribute to rebound, 29.7\%
to suppression, and 43.7\% remain functionally neutral. This suggests that more robust negation
following might be achievable through precision modifications rather than holistic architectural
changes. See Appendix \ref{sec: mech interp} for further results.
\FloatBarrier

\section{Conclusion}
LLMs display ironic rebound: after a “do not mention X” instruction, they become more likely to output X, with circuit tracing showing a sparse set of middle-layer attention heads that amplify the forbidden token under load even as early layers suppress it. Semantic distractors induce the strongest rebound, while simple repetition suppresses the most. Models with sharper polarity separation show longer-lasting rebound, reflected in higher L50 values. Despite limits such as a synthetic English-only dataset, log-probability evaluations, and a small set of distractor types, the results expose the fragility of negation-based filters and explain why this fragility can persist across context lengths, since sparse amplification heads can overpower early suppression when load and framing align, motivating tests in natural dialogue and with multi-word targets.

\bibliographystyle{plain}
\bibliography{citations}

\newpage
\appendix
\section{Appendix}
\setcounter{figure}{0}

\subsection{Dataset Details}
\label{sec: dataset details}

\subsubsection*{Sources}
ReboundBench combines and extends two public negation datasets:
\begin{itemize}
    \item \textbf{This-is-not-a-dataset} (HiTZ)\citep{anschütz2023correct}
    \item \textbf{NUBench} (SNU Thunder)\citep{so2025thundernubenchbenchmarkllmssentencelevel}
\end{itemize}

\subsubsection*{Preparation}
We extracted English examples, selected common nouns that tokenize as a single token, and normalized text (lowercasing, removing punctuation). Each item was wrapped in a template of the form:

\begin{quote}
\texttt{[Topic]. Do not mention [Target]. [Distractor text] ...}
\end{quote}

Distractors varied by \emph{type} (semantic, syntactic, repetition) and \emph{length} (0–1024 tokens). For polarity discrimination, each (topic, target) pair was expanded into negative/neutral continuations. The final dataset contains about 5{,}000 items in CSV format with fields: topic, target, distractor type, distractor length, and prompt.

\subsubsection*{Examples}
\begin{itemize}
    \item Topic: Paris travel; Target: \texttt{eiffel}; Repetition distractor:
    Write about a trip to Paris. Do not mention the Eiffel. Paris is a place many tourists visit.''
    \item Topic: Farm animals; Target: \texttt{sheep}; Semantic distractor:
    Write about animals on a farm. Do not mention sheep. Cows and pigs are often raised for food.''
\end{itemize}

\subsubsection*{Release}
We provide the dataset and scripts for generation as a CSV file with $\sim$5k prompts for reproducibility and further study.

\newpage
\section{Future Work/Limitations}
\label{sec: future work/limitation}
Several limitations constrain the generalizability of our findings. First, our dataset consists of synthetic, templated prompts rather than naturalistic conversational contexts, potentially limiting ecological validity. The reliance on log-probabilities rather than actual text generation may not fully capture how rebound manifests in practical deployment scenarios where models produce complete responses. Additionally, our analysis focused exclusively on English and single-token target concepts, leaving open questions about cross-linguistic generalization and multi-word forbidden phrases.
Our experimental design, while systematic, employed only three distractor types (semantic, syntactic, repetition) and may not capture the full spectrum of cognitive load conditions that occur in real-world usage. The artificial nature of our "do not mention X" instructions also differs from the more nuanced content policies and safety filters deployed in production systems, which often involve complex multi-step reasoning about appropriateness rather than simple lexical prohibition.

Future research should address several critical directions. First, mechanistic interpretability work is needed to identify the specific circuits responsible for ironic rebound, potentially building on existing analyses of copy-suppression heads to understand how negation instructions are processed and how they interact with attention mechanisms across different context lengths. Second, studies should explore more naturalistic settings, including conversational contexts where negation emerges organically rather than through explicit instructions.

The anomalous behavior of GPT-OSS-20B, which contradicts the observed scaling trends, warrants further investigation. We hypothesize this may stem from its specific fine-tuning process, which was based on a larger proportion of synthetic data than the other models we observed, or from architectural variations not present in other models of similar scale.

Investigation of mitigation strategies represents another crucial avenue. If negation-based filters are inherently fragile, alternative approaches such as positive framing ("focus on Y instead of X") or multi-step reasoning chains may prove more robust. Additionally, the scaling trends we observed suggest that architectural modifications or training procedures might be developed to reduce rebound susceptibility while preserving the enhanced polarity discrimination capabilities of larger models.
Finally, direct behavioral comparisons between human and model rebound patterns could clarify whether observed effects reflect genuine cognitive parallels or merely superficial statistical regularities, informing both our understanding of language model cognition and the development of more human-aligned AI systems.

\subsection{Regression Results}
\FloatBarrier
\begin{table}[!htbp]
    \centering
    \begin{tabular}{lcccccc}
    \toprule
    \textbf{Coefficient} & \textbf{Coef.} & \textbf{Std.Err.} & \textbf{P>|z|} & \textbf{[0.025, 0.975]} \\
    \midrule
        GPT-NeoX-20B & 0.366 & 0.051 & 0.000 & [0.266, 0.466] \\
        GPT-OSS-20B & -10.692 & 0.051 & 0.000 & [-10.792, -10.592] \\
        GPT-2-Small & -3.446 & 0.051 & 0.000 & [-3.546, -3.346] \\
        LFM2-350M & -3.995 & 0.051 & 0.000 & [-4.095, -3.895] \\
        Llama-3-8B-Instruct & -4.822 & 0.051 & 0.000 & [-4.922, -4.722] \\
        OPT-2.7B & -2.398 & 0.051 & 0.000 & [-2.498, -2.298] \\
        Pythia-410M & -4.789 & 0.051 & 0.000 & [-4.888, -4.689] \\
        Qwen3-14B & -1.643 & 0.051 & 0.000 & [-1.743, -1.544] \\
    \bottomrule
    \end{tabular}
    \caption{Mixed-effects regression results for the surprisal difference $\Delta$(bits).}
    \label{tab:regression}
\end{table}

\begin{table}[!htbp]
\centering

\begin{tabular}{lcccccc}
    \toprule
    \textbf{Coefficient} & \textbf{Coef.} & \textbf{Std.Err.} & \textbf{P>|z|} & \textbf{[0.025, 0.975]} \\
    \midrule
        GPT-NeoX-20B & 0.3661 & 0.469 & 0.435 & [-0.553, 1.286] \\
        GPT-OSS-20B & -10.6916 & 0.521 & 0.000 & [-11.712, -9.671] \\
        GPT-2-Small & -3.4459 & 0.518 & 0.000 & [-4.460, -2.431] \\
        LFM2-350M & -3.9949 & 0.449 & 0.000 & [-4.876, -3.114] \\
        Llama-3-8B-Instruct & -4.8224 & 0.569 & 0.000 & [-5.938, -3.707] \\
        OPT-2.7B & -2.3977 & 0.473 & 0.000 & [-3.324, -1.471] \\
        Pythia-410M & -4.7886 & 0.456 & 0.000 & [-5.682, -3.895] \\
        Qwen3-14B & -1.6435 & 0.458 & 0.000 & [-2.541, -0.746] \\
    \bottomrule
    \end{tabular}
    \caption{OLS Regression Results for the surprisal difference $\Delta$(bits).}
    \label{tab:ols_regression}
\end{table}

\textit{Notes: R-squared: 0.564, Adjusted R-squared: 0.564, F-statistic: 2563, Prob (F-statistic): 4.79e-86. Standard errors are robust to cluster correlation.}

\FloatBarrier
\subsection{Detailed Results: Ironic Rebound vs. Load}

\setcounter{figure}{0}
\renewcommand{\thefigure}{B2.\arabic{figure}}

\begin{figure}[htbp]
    \centering
    \begin{minipage}[t]{0.32\textwidth}
        \centering
        \includegraphics[width=\textwidth]{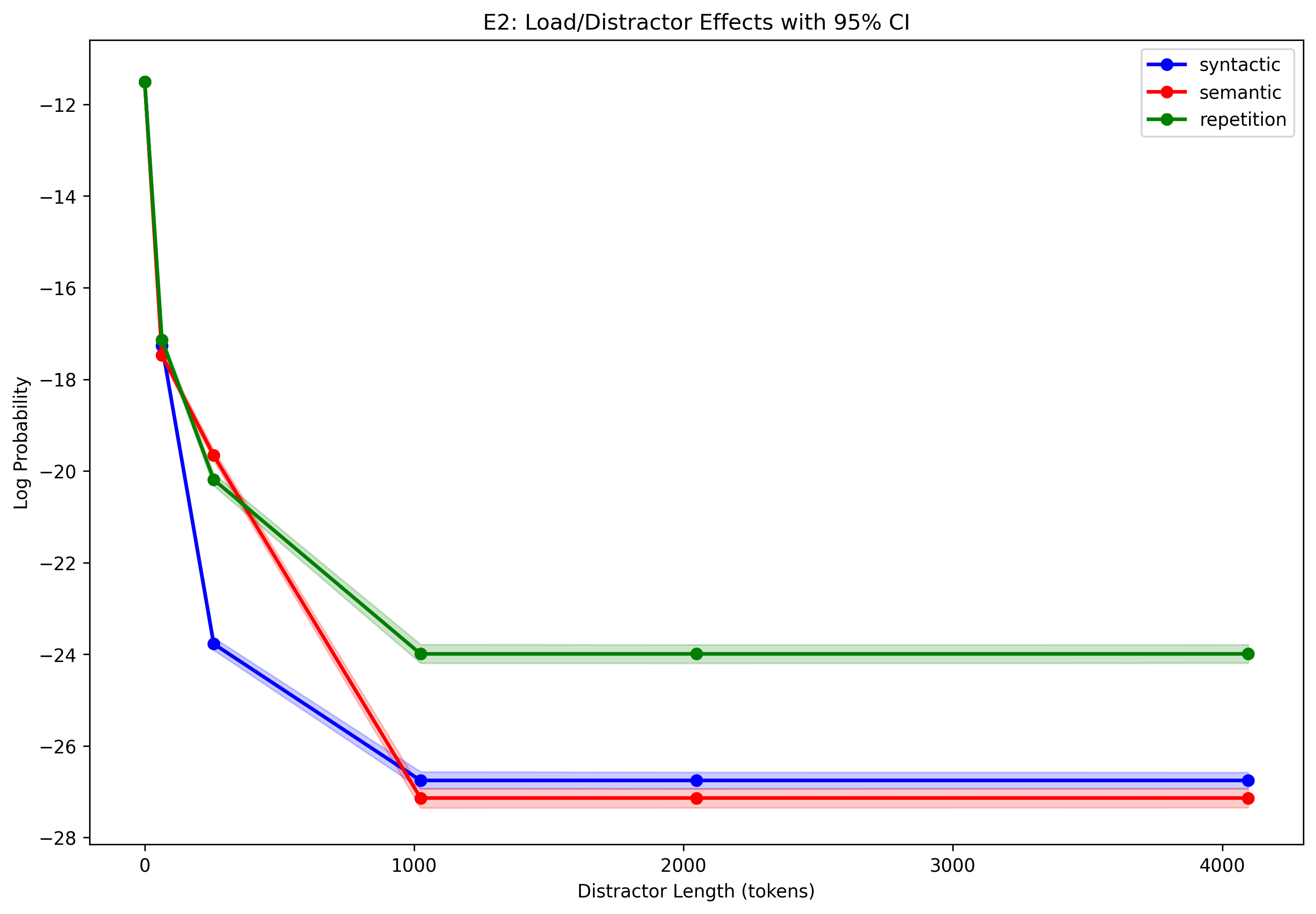}
        \caption{GPT-2-Small}
        \label{fig:gpt2-small}
    \end{minipage}\hfill
    \begin{minipage}[t]{0.32\textwidth}
        \centering
        \includegraphics[width=\textwidth]{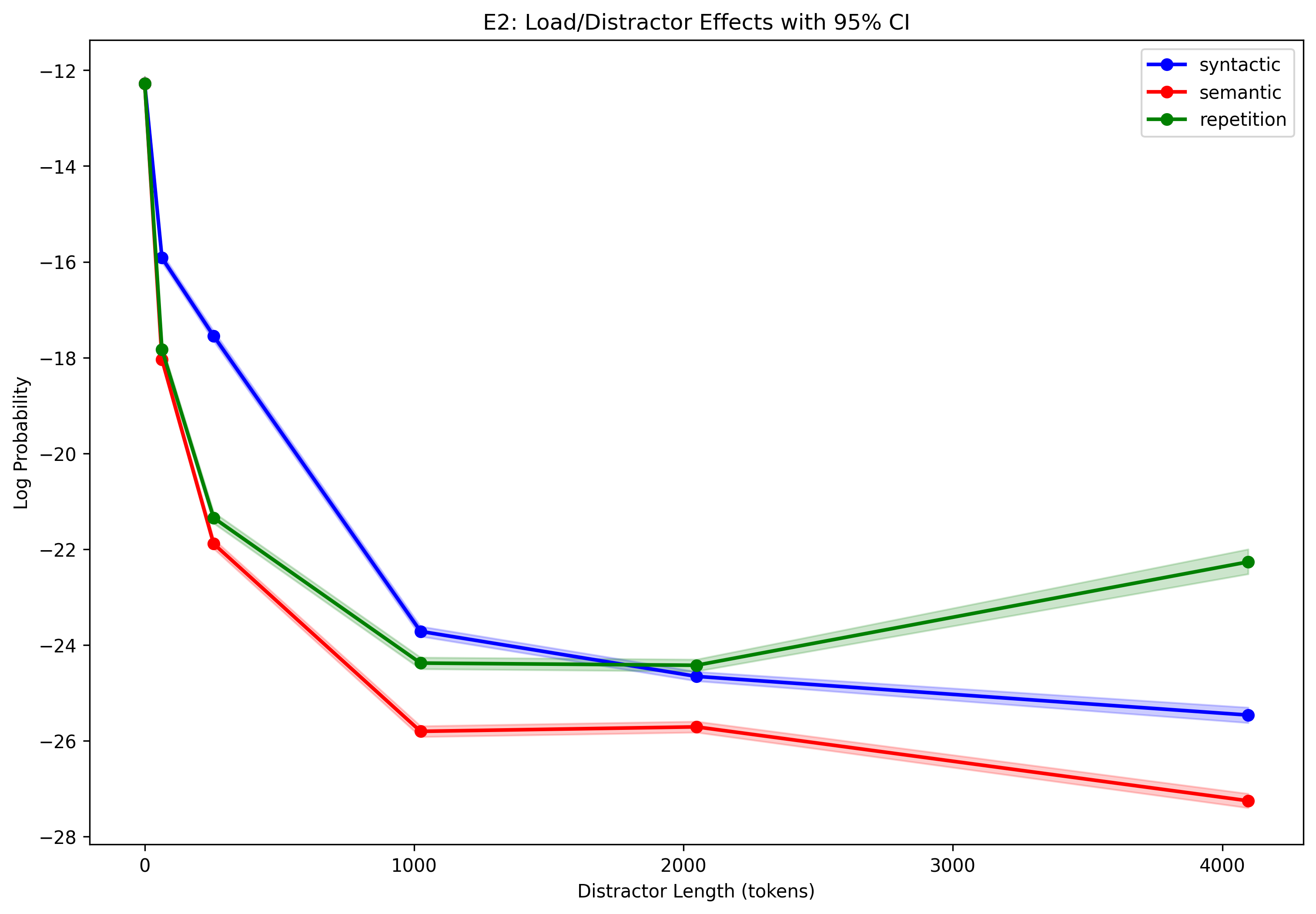}
        \caption{LFM2-350M}
        \label{fig:lfm2-350m}
    \end{minipage}\hfill
    \begin{minipage}[t]{0.32\textwidth}
        \centering
        \includegraphics[width=\textwidth]{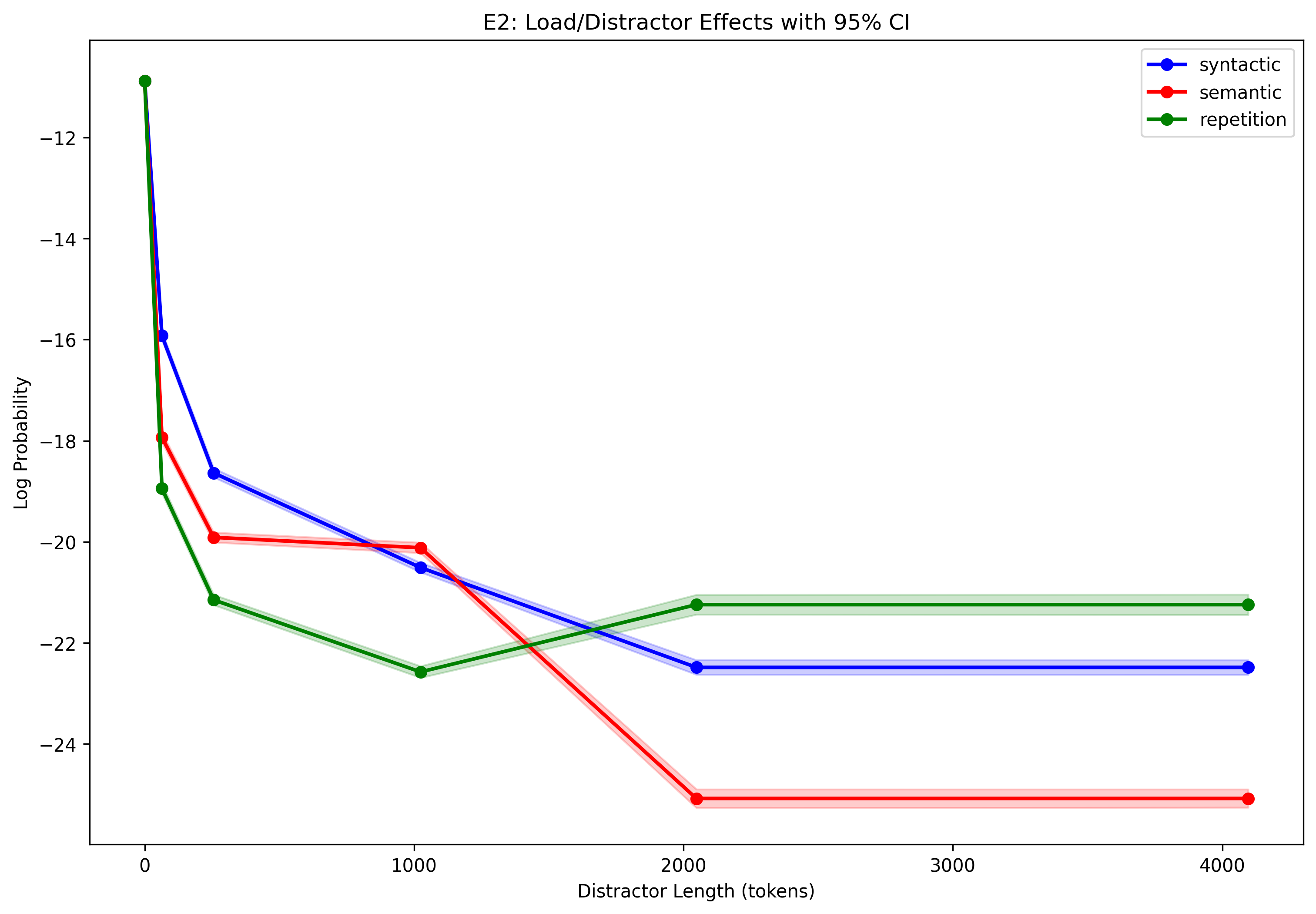}
        \caption{Pythia-410M}
        \label{fig:pythia-410m}
    \end{minipage}
\end{figure}

\FloatBarrier
\begin{figure}[htbp]
    \centering
    \begin{minipage}[t]{0.32\textwidth}
        \centering
        \includegraphics[width=\textwidth]{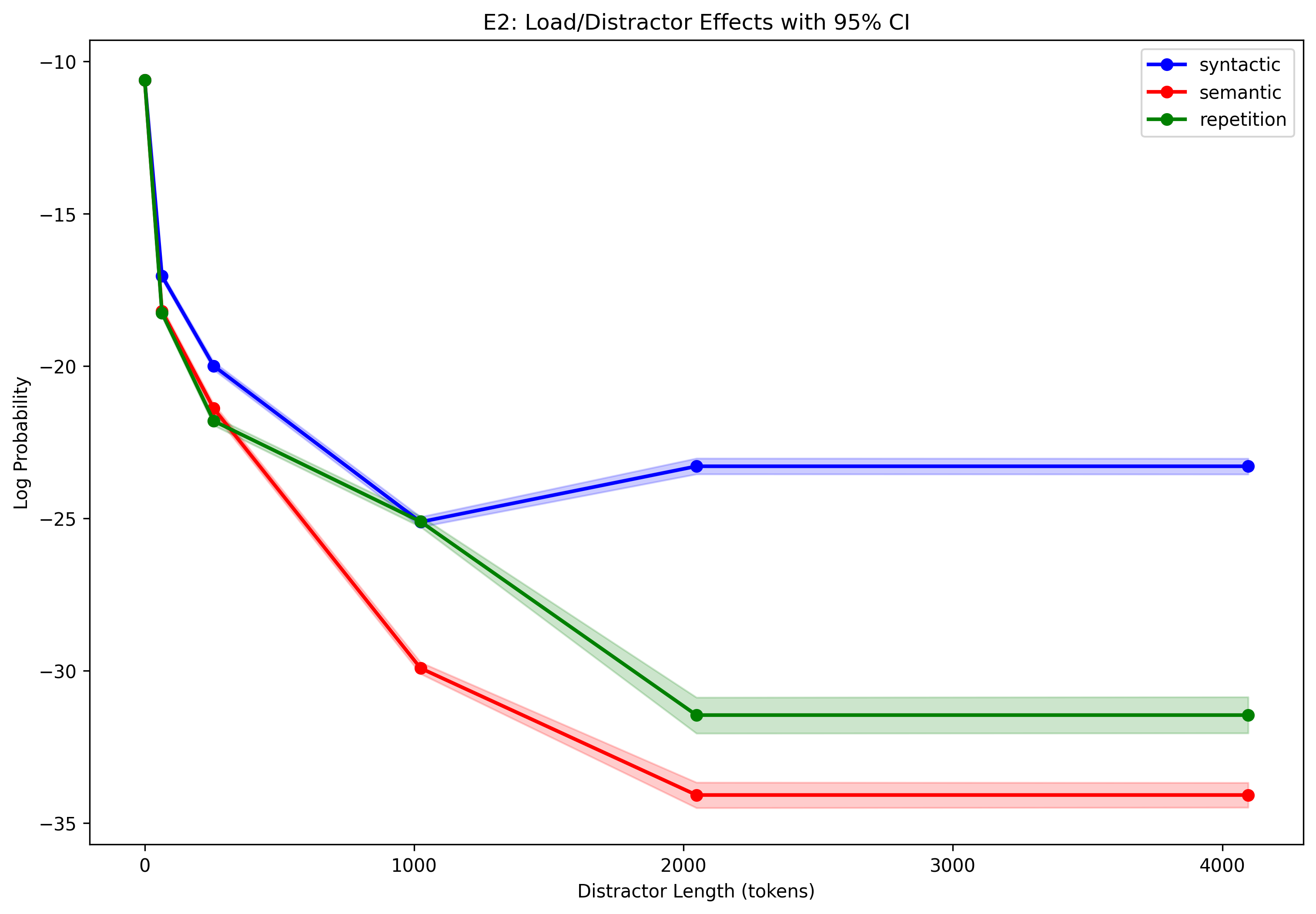}
        \caption{Bloom-560M}
        \label{fig:bloom-560m}
    \end{minipage}\hfill
    \begin{minipage}[t]{0.32\textwidth}
        \centering
        \includegraphics[width=\textwidth]{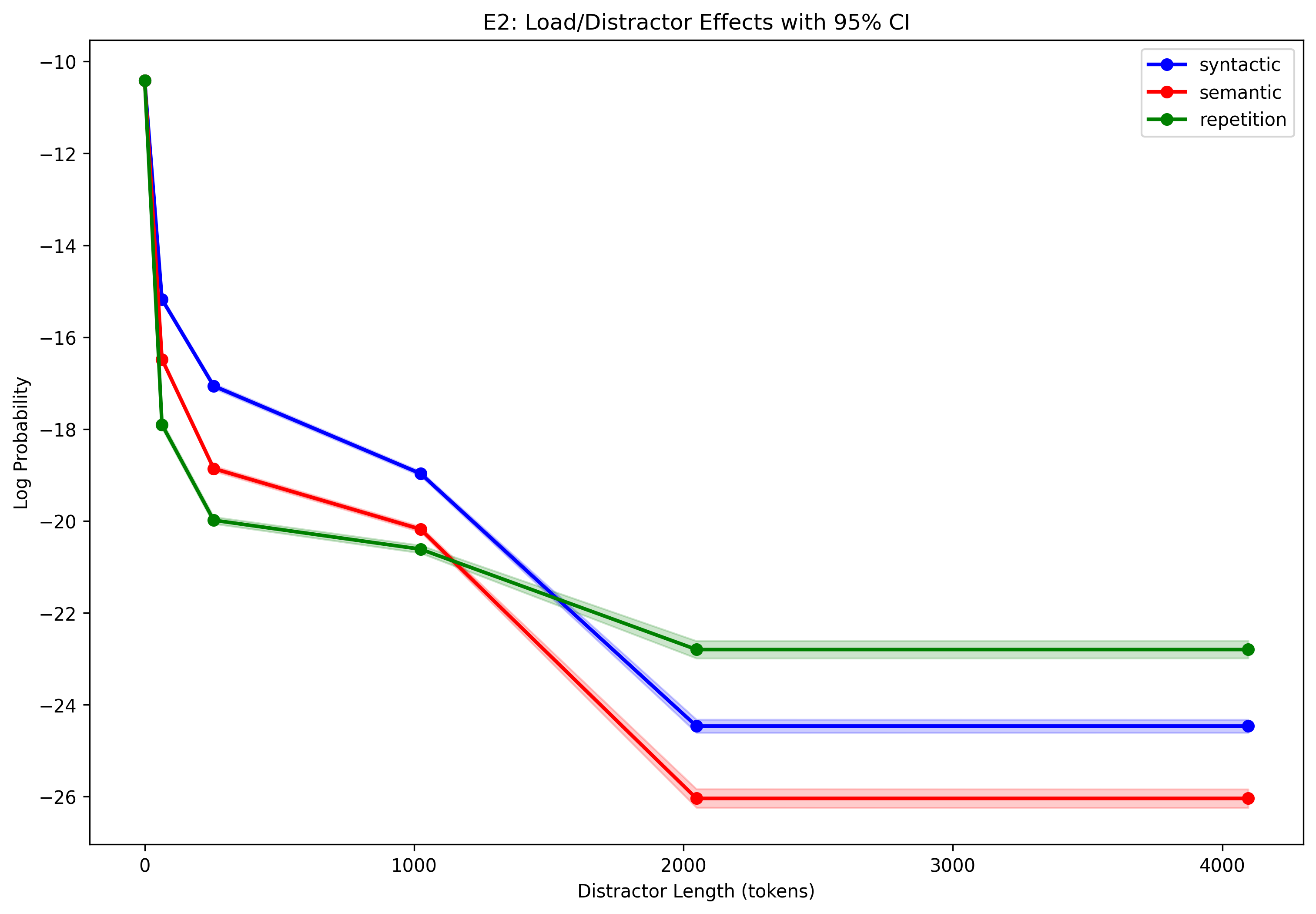}
        \caption{OPT-2.7B}
        \label{fig:opt-2.7b}
    \end{minipage}\hfill
    \begin{minipage}[t]{0.32\textwidth}
        \centering
        \includegraphics[width=\textwidth]{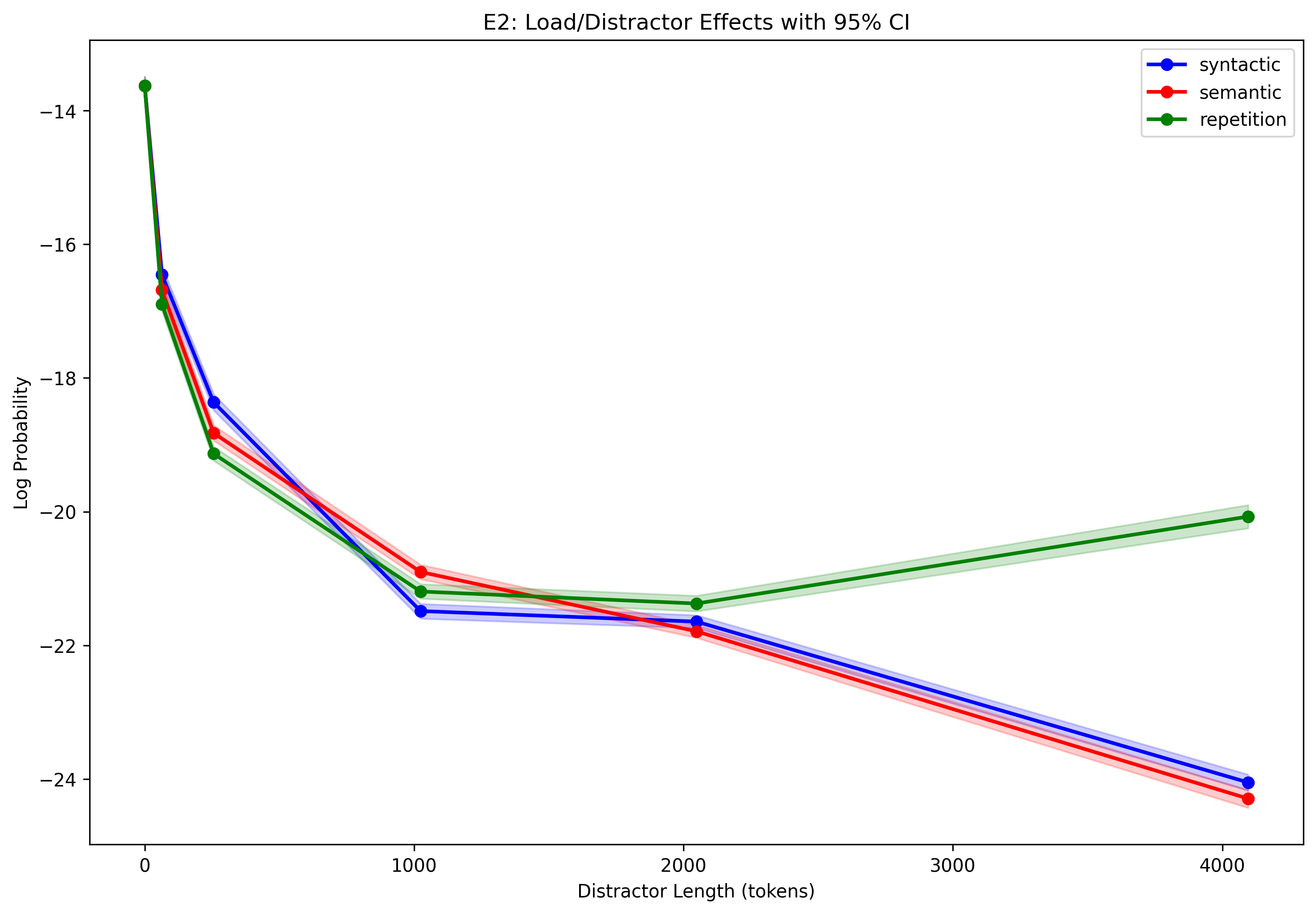}
        \caption{\shortstack{Llama-3-8B-\\[-0.2em]Instruct}}
        \label{fig:llama-3-8b-instruct}
    \end{minipage}
\end{figure}

\begin{figure}[htbp!]
    \centering
    \begin{minipage}[t]{0.32\textwidth}
        \centering
        \includegraphics[width=\textwidth]{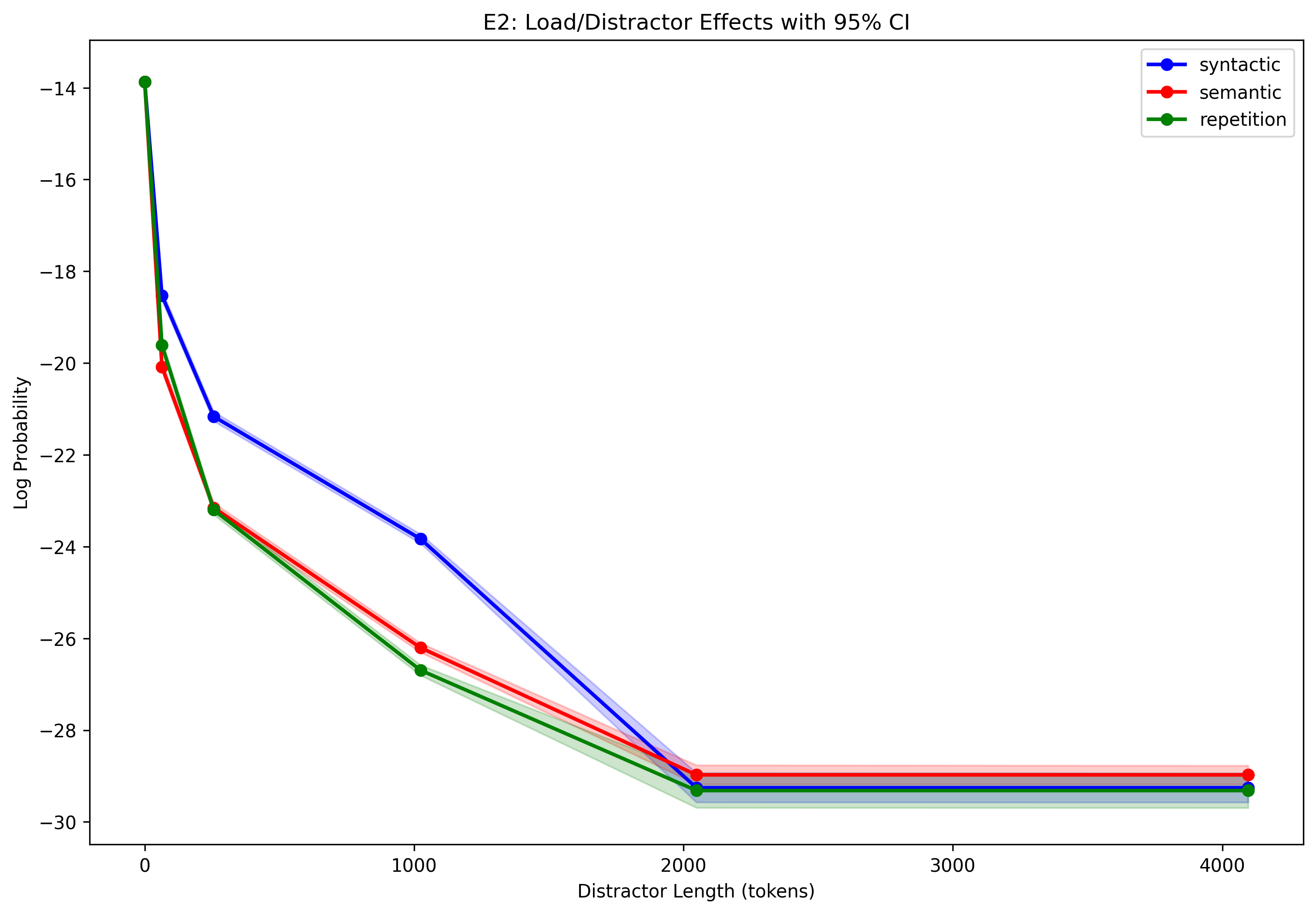}
        \caption{Qwen3-14B}
        \label{fig:qwen3-14b}
    \end{minipage}\hfill
    \begin{minipage}[t]{0.32\textwidth}
        \centering
        \includegraphics[width=\textwidth]{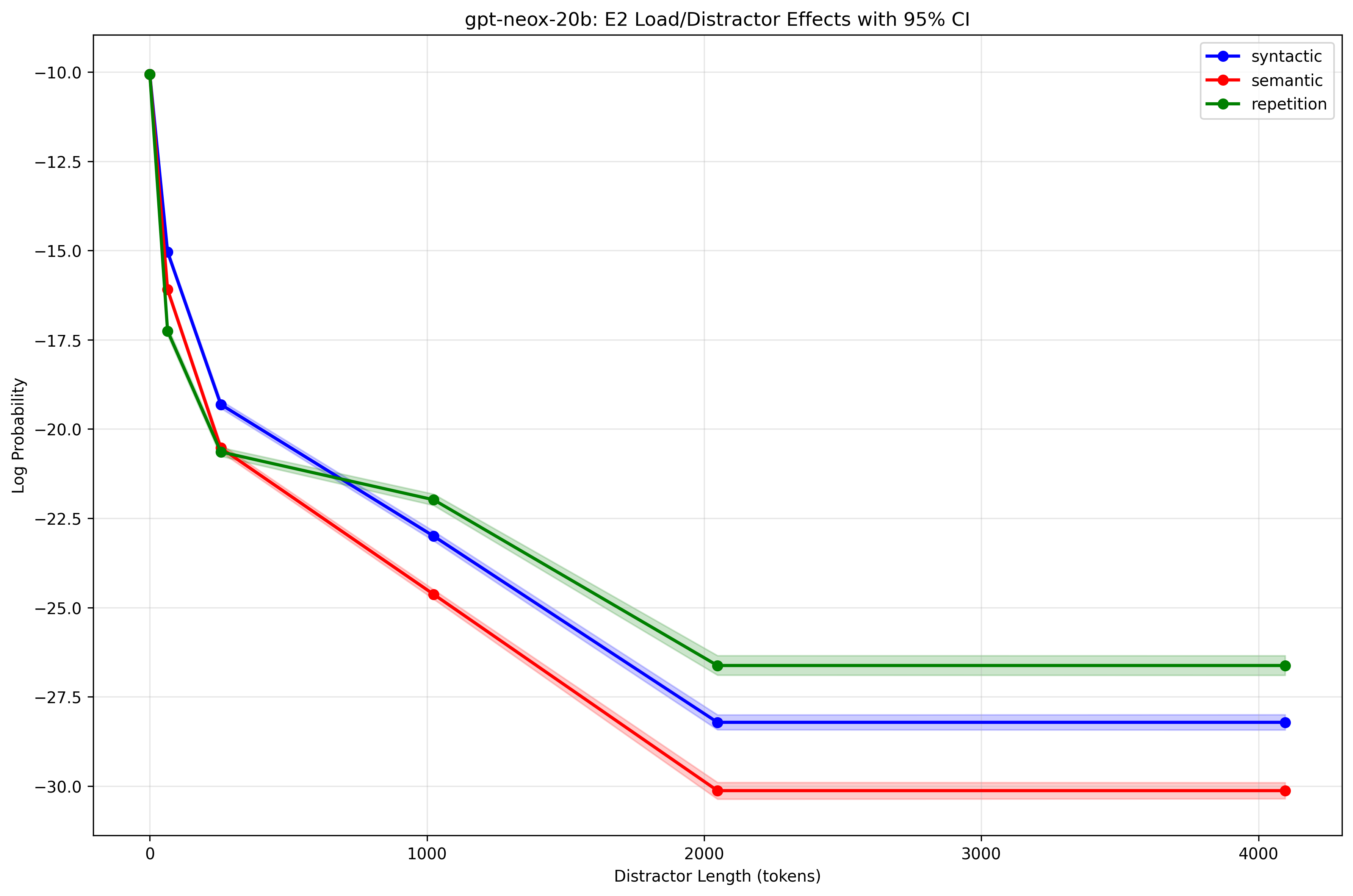}
        \caption{GPT-NeoX-20B}
        \label{fig:gpt-neox-20b}
    \end{minipage}\hfill
    \begin{minipage}[t]{0.32\textwidth}
        \centering
        \includegraphics[width=\textwidth]{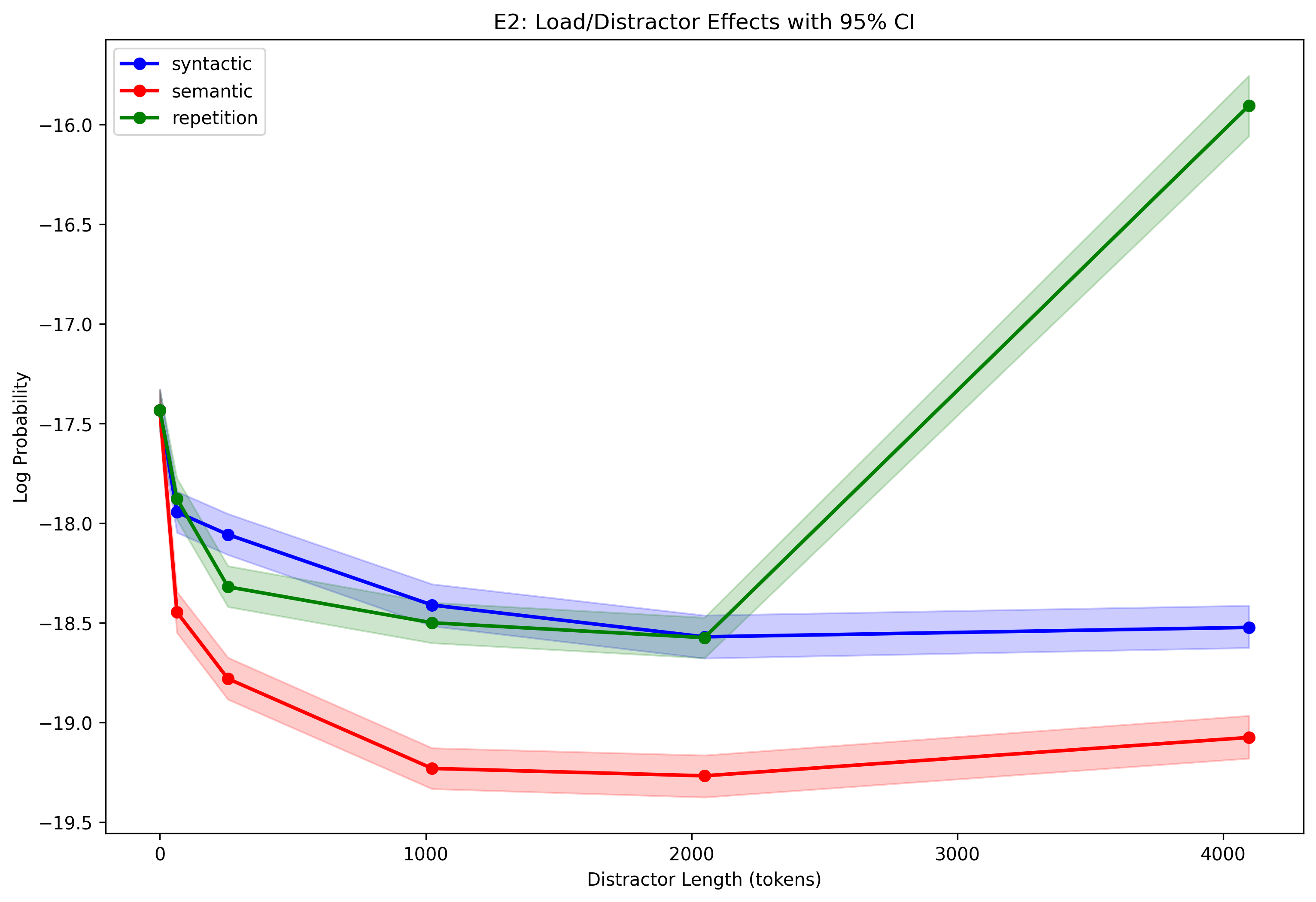}
        \caption{GPT-OSS-20B}
        \label{fig:gpt-oss-20b}
    \end{minipage}
\end{figure}
\FloatBarrier
\label{fig:appendix-model-graphs}

\newpage
\begin{table}[htbp!]
\centering
\begin{tabular}{l l c c c}
  \toprule
  Model & Distractor Type & Peak $\Delta$ (bits) & $L_{50}$ & AUC$\Delta$ \\
  \midrule
  GPT-2 Small          & Semantic   & 18.2  & 0.09 & 1.73 \\
                       & Syntactic  & 17.8  & 0.05 & 0.91 \\
                       & Repetition & 13.9  & 0.06 & 0.98 \\
  \midrule
  LFM2-350M            & Semantic   & 17.4  & 0.06 & 2.09 \\
                       & Syntactic  & 16.4  & 0.14 & 2.49 \\
                       & Repetition & 10.4  & 0.04 & 0.38 \\
  \midrule
  Pythia-410M          & Semantic   & 15.4  & 0.06 & 2.89 \\
                       & Syntactic  & 13.1  & 0.06 & 1.62 \\
                       & Repetition & 9.1  & 0.04 & 0.21 \\
  \midrule
  Bloom-560M           & Semantic   & 28.4  & 0.13 & 4.26 \\
                       & Syntactic  & 13.7  & 0.05 & 0.88 \\
                       & Repetition & 24.6  & 0.13 & 4.35 \\
  \midrule
  OPT-2.7B             & Semantic   & 18.2  & 0.20 & 3.61 \\
                       & Syntactic  & 16.8  & 0.23 & 3.50 \\
                       & Repetition & 12.5  & 0.05 & 1.49 \\
  \midrule
  Llama-3-8B-Instruct  & Semantic   & 13.2  & 0.15 & 2.83 \\
                       & Syntactic  & 13.0  & 0.14 & 2.58 \\
                       & Repetition & 6.9  & 0.05 & 0.31 \\
  \midrule
  Qwen3-14B            & Semantic   & 17.3  & 0.06 & 2.33 \\
                       & Syntactic  & 18.8  & 0.18 & 3.64 \\
                       & Repetition & 18.1  & 0.06 & 2.35 \\
  \midrule
  GPT-NeoX-20B         & Semantic   & 24.6  & 0.12 & 4.06 \\
                       & Syntactic  & 22.6  & 0.12 & 3.80 \\
                       & Repetition & 18.7  & 0.06 & 3.00 \\
  \midrule
  GPT-OSS-20B          & Semantic   & 1.6  & 0.05 & 0.09 \\
                       & Syntactic  & 1.2  & 0.09 & 0.15 \\
                       & Repetition & -2.5  & 0.95 & 0.00 \\
  \bottomrule
  \end{tabular}
\caption{Summary of rebound dynamics across models and distractor types.
Peak $\Delta$ indicates the maximum rebound in surprisal bits (higher = stronger rebound).
$L_{50}$ is the normalized load at which rebound falls to half of its peak (higher = more persistent).}
\label{tab:rebound_summary}
\end{table}

\newpage
\section{Further Attention Head Analysis}
\label{sec: mech interp}
\setcounter{figure}{0}
\renewcommand{\thefigure}{C1.\arabic{figure}}

\subsection{Most Attention Heads Have Little Effect on Rebound}

\begin{figure}[!htbp]
    \centering
    \includegraphics[width=0.85\textwidth]{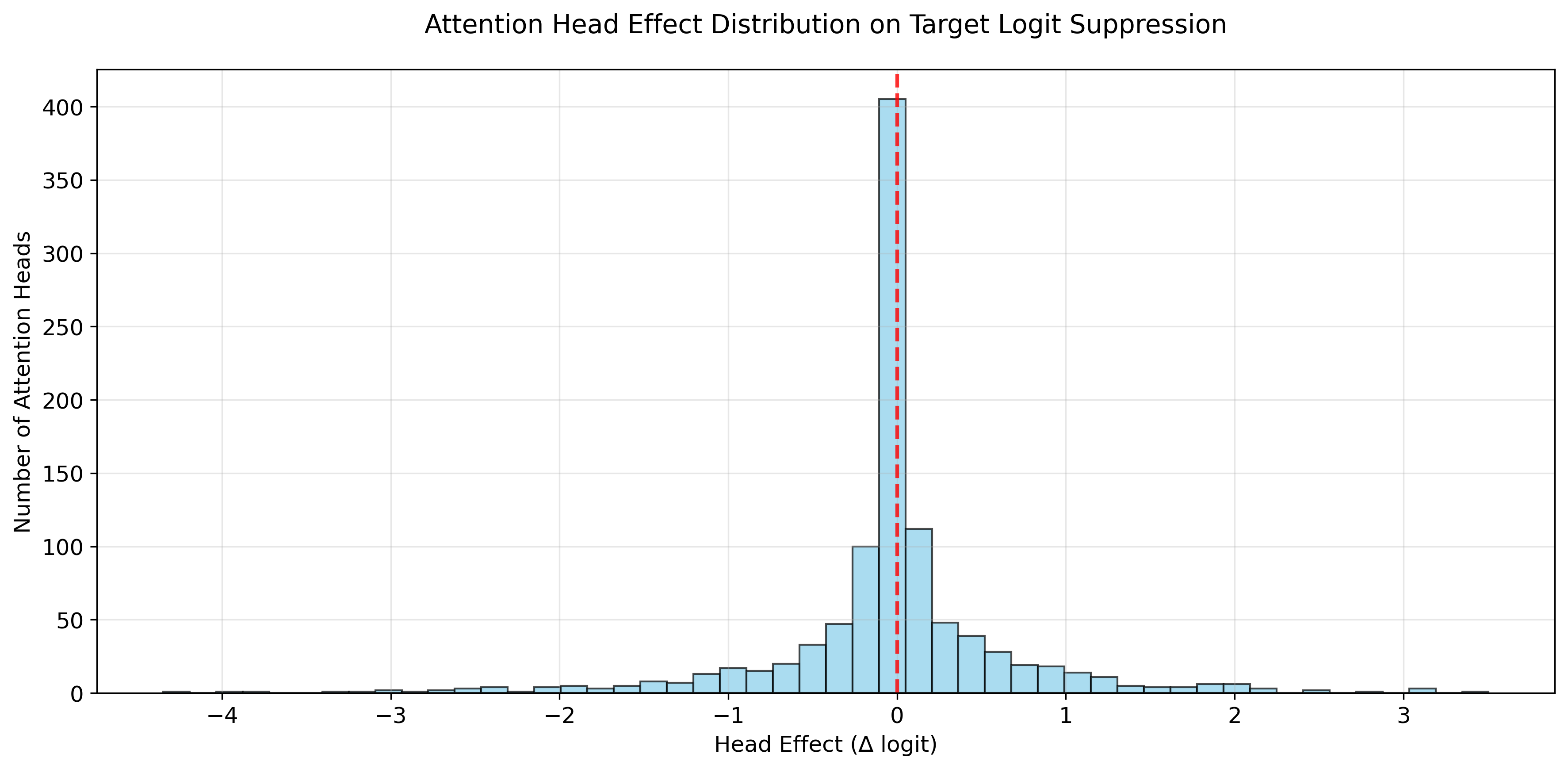}
    \caption{Distribution of attention head effects on forbidden token probability. Most heads have minimal impact, but a sparse few create strong suppression or amplification effects.}
    \label{fig:head_distribution}
\end{figure}
\FloatBarrier

We tested all attention heads in Llama 3 8B Instruct, a 32-layer model (over 1{,}000 heads total) that showed a clear rebound effect in our initial measurement. We found that the vast majority of heads had essentially no impact on whether the model produces forbidden tokens. However, a small number of heads had large effects, either strongly suppressing forbidden words (negative effects) or amplifying them (positive effects). Figure \ref{fig:head_distribution} shows this distribution, with most heads clustered near zero but a few outliers reaching extreme values of up to 4 logits in either direction.

This pattern suggests that ironic rebound emerges from the interaction of a relatively small number of specialized components, rather than being distributed across the entire model.

\subsection{Summary of Internal Mechanisms}

Our circuit analysis reveals that ironic rebound is not simply a failure of language models to follow instructions. Instead, it emerges from identifiable internal mechanisms: early layers implement suppression as intended, but a sparse set of amplifying heads in middle layers partially reverses these effects. Across the full model architecture, 26.7\% of heads contribute to rebound, 29.7\% to suppression, and 43.7\% remain functionally neutral. This suggests that more robust negation following might be achievable through precision modifications rather than holistic architectural changes. 

Moreover, the dynamics we observe raise questions about whether different types of distractors have varying effects on rebound. Semantic distractors may engage the same middle-layer heads responsible for amplification, while repetitive text might strengthen the early suppression circuits. Understanding these internal mechanisms opens new avenues for both explaining and potentially controlling ironic rebound in future language models.

\end{document}